\documentclass{article} 
\usepackage{nips15submit_e,times}
\usepackage{hyperref}
\usepackage{url}
\usepackage{amsmath}
\usepackage{placeins}
\usepackage{graphicx}
\usepackage{float}
\usepackage[usestackEOL]{stackengine}
\setstackgap{L}{\normalbaselineskip}
\usepackage{enumitem}
\usepackage{multirow}
\usepackage{array}

\title{Political Speech Generation}
\author{
Valentin Kassarnig\\
College of Information and Computer Sciences\\
University of Massachusetts Amherst\\
vkassarnig@umass.edu 
}
\nipsfinalcopy 

\usepackage{tabularx}
\usepackage{hyperref}

\begin{document}
\maketitle

\begin{abstract}
In this report we present a system that can generate political speeches for a desired political party. Furthermore, the system allows to specify whether a speech should hold a supportive or opposing opinion. The system relies on a combination of several state-of-the-art NLP methods which are discussed in this report. These include n-grams, Justeson \& Katz POS tag filter, recurrent neural networks, and latent Dirichlet allocation. Sequences of words are generated based on probabilities obtained from two underlying models: A language model takes care of the grammatical correctness while a topic model aims for textual consistency. Both models were trained on the Convote dataset which contains transcripts from US congressional floor debates. Furthermore, we present a manual and an automated approach to evaluate the quality of generated speeches. In an experimental evaluation generated speeches have shown very high quality in terms of grammatical correctness and sentence transitions.
\end{abstract}

\section{Introduction}
Many political speeches show the same structures and same characteristics regardless of the actual topic. Some phrases and arguments appear again and again and indicate a certain political affiliation or opinion. We want to use these remarkable patterns to train a system that generates new speeches. Since there are major differences between the political parties we want the system to consider the political affiliation and the opinion of the intended speaker. The goal is to generate speeches where no one can tell the difference to hand-written speeches. 

In this report we first discuss related works which deal with similar or related methods. Then we describe and analyze the dataset we use. Next, we present the methods we used to implement our system. We also describe investigated methods that were not used in the final implementation. Then we describe a performed experiment and how we evaluated the results. Finally, we conclude our work and give an outlook. The appendix of this report contains the generated speeches from the experiment.

\section{Related work}
\label{sec:related}
Creating models for a corpus that allow retrieving certain information is a major part of this project as well as in the entire NLP domain.  Blei et al.~\ref{itm:ref1} present in their paper a model which is known as latent Dirichlet allocation (LDA). LDA has become one of the most popular topic models in the NLP domain. LDA is generative probabilistic model that discovers automatically the underlying topics. Each document is modeled as a mixture of various topics. These topics can be understood as a collection of words that have different probabilities of appearance. Words with the highest probabilities represent the topics.

However, LDA is a bag-of-words model which means that the word orders are not preserved. That means LDA does not capture collocations or multiword named entities. Lau et al.~\ref{itm:ref2} claim that collocations empirically enhance topic models. In an experiment they replaced the top-ranked bigrams with single tokens, deleted the 200 most frequent terms from the vocabulary and performed ordinary LDA. The results from experiments on four distinct datasets have shown that this bigram-variant is very beneficial for LDA topic models.

Fürnkranz~\ref{itm:ref3} has studied the usage of n-grams in the text-categorization domain. He has shown that using bi- and trigrams in addition to the set-of-word representation improves the classification performance significantly. Furthermore, he has shown that sequences longer than three words reduce the classification performance. That also indicates that collocations play a crucial role when it comes to inferring the latent structure of documents.

Cavnar and Trenkle~\ref{itm:ref4} have also used an n-gram-based approach for text categorization. Their system is based on calculating and comparing profiles of N-gram frequencies. They compute for every category a representing profile from the training data. Then the system computes a profile for a particular document that is to be classified. Finally, the system computes a distance measure between the document's profile and each of the category profiles and selects the category whose profile has the smallest distance.

Smadja~\ref{itm:ref5} presents a tool, \textit{Xtract}, which implements methods to extracts variable-length collocations. The extraction process is done in several stages. In the first stage the system determines the top-ranked bigrams of the corpus. In the second stage \textit{Xtract} examines the statistical distribution of words and part-of-speech tags around the bigrams from the previous stage. Compounds with a probability above a certain threshold are retained while the others are rejected. In the third stage they enrich the collocations with syntactical information obtained from Cass~\ref{itm:ref6}. The syntactical information helps to evaluate the candidate collocations and to decide whether they should be rejected or not. 

Wang et al~\ref{itm:ref7} propose a topical n-gram model that is capable of extracting meaningful phrases and topics. It combines the bigram topic model~\ref{itm:ref8} and LDA collocation model~\ref{itm:ref9}. One of the key features of this model is to decide whether two consecutive words should be treated as a single token or not depending on their nearby context. Compared to LDA the extracted topics are semantically more meaningful. This model shows also really good results in information retrieval (IR) tasks.

Justeson and Katz~\ref{itm:ref10} present a method to extract technical terms from documents. Their approach is not restricted to technical terms but applies to all multiword named entities of length two or three. The foundations of their method are bi- and trigrams which have a certain POS tag structure. That is, they extract all bi- and trigrams from the corpus, identify their POS tags and check them against a predefined list of accepted POS tag patterns. In their experiment this method identifies 99\% of the technical multiword terms in the test data.

Wacholder~\ref{itm:ref11} presents an approach for identifying significant topics within a document. The proposed method bases on the identification of Noun Phrases (NPs) and consists of three steps. First, a list of candidate significant topics consisting of all simplex NPs is extracted from the document. Next, these NPs are clustered by head.  Finally, a significance measure is obtained by ranking frequency of heads. Those NPs with heads that occur with greater frequency in the document are more significant than NPs whose head occurs less frequently.

Blei and Lafferty~\ref{itm:ref12} propose their Correlated Topic model (CTM). While LDA assumes all latent topics are independent CTM aims to capture correlations between them. They argue that a document about genetics is more likely also about disease than X-ray astronomy. The CTM builds on the LDA model but they use a hierarchical topic model of documents that replaces the Dirichlet distribution of per-document topic proportions with a logistic normal. According to their results the model gives better predictive performance and uncovers interesting descriptive statistics.

Ivyer et al.~\ref{itm:ref19} apply Recursive Neural Networks (RNN) to political ideology detection. The RNNs were initialized with word2vec embeddings. The word vector dimensions were set to 300 to allow direct comparison with other experiments. However, they claim that smaller vector sizes (50, 100) do not significantly change accuracy. They performed experiments on two different dataset: the Convote dataset~\ref{itm:ref25} and the Ideological Books Corpus (IBC)~\ref{itm:ref21}. They claim that their model outperforms existing models on these two datasets. 

There has been a lot of research in the field of Natural Language Generation (NLG). The paper \textit{Building Applied Natural Language Generation Systems}~\ref{itm:ref13} discusses the main requirements and tasks of NLG systems. Among others, they investigate a so-called Corpus-based approach. That is, a collection of example inputs is mapped to output texts of the corpus. This is basically what we plan to do because we have already all the speech segments labeled with the political party and the opinion. However, our generator will have a simpler architecture but we will use the described list of tasks as a guideline.

Most NLG systems are designed to create a textual representation of some input data. That is, the input data determines the content. For example SumTime-Mousam~\ref{itm:ref14} generates a textual weather forecast based on numerical weather simulations. Another example is the ModelExplainer system~\ref{itm:ref15} which takes as input a specification of an object-oriented class model and produces as output a text describing the model. Other NLG systems are used as authoring aid for example to help personnel officers to write job descriptions~\ref{itm:ref16} or to help technical authors produce instructions for using software~\ref{itm:ref17}.

A NLG system that follows a different approach is SciGen~\ref{itm:ref22}. SciGen is an automatic computer science research paper generator developed by three MIT students. That is, it creates random papers which show actually a very high quality in terms of structuring and lexicalization, and they even include graphs, figures, and citations.  SciGen has become pretty famous after some of its generated papers got accepted at conferences and published in journals. In particular, their paper \textit{Rooter: A Methodology for the Typical Unification of Access Points and Redundancy} raised a lot of attention because it was accepted to the \textit{2005 World Multiconference on Systemics, Cybernetics and Informatics} (WMSCI) and the authors were even invited to speak at the conference. SciGen requires as input only the names of the authors; all the content will be generated randomly. Our generator will follow the same approach since we also do not specify the content of the generated speech. The content is determined by the training data and requires no further specification.

\section{Data set}
\label{sec:dataset}
The main data source for this project is the Convote data set~\ref{itm:ref25}. It contains a total of 3857 speech segments from 53 US Congressional floor debates from the year 2005. Each speech segment can be referred to its debate, its speaker, the speaker's party and the speaker's vote which serves as the ground-truth label for the speech. The dataset was originally created in the course of the project \textit{Get out the vote}~\ref{itm:ref18}. The authors used the dataset to train a classifier in order to determine whether a speech represents support of or opposition to proposed legislation. They did not only analyze the speeches individually but also investigated agreements and disagreements with the opinions of other speakers. That is, they identified references in the speech segments, determined the targets of those references, and decided whether a reference represents an instance of agreement or disagreement. However, we focus only on the individual speech segments and disregard references.

For our work we have removed single-sentence speeches, HTML-tags and corrected punctuation marks. In order to enable simple sentence splitting we replaced all sentence delimiters by a stop-token. Furthermore, we inserted special tokens which indicate the start and the end of a speech. Then we divided all the speeches into the four classes given by the combination of possible political parties and speech opinions. Table~\ref{tab:table1} shows the four speech classes and table~\ref{tab:table2} gives a quantitative overview of the corpus' content. It can be seen that the classes \textit{RY} and \textit{DN} contain the majority of the speeches.

\begin{table}
\centering
\begin{tabular}{ | c | c | c | c | }
  \cline{3-4}
  \multicolumn{2}{c|}{ } & \multicolumn{2}{c|}{\textbf{Party}}\\ \cline{3-4}
  \multicolumn{2}{c|}{ }& Republicans & Democrats\\ \hline
  \multirow{2}{*}{\textbf{Opinion}} & Support (Yea) & \textit{RY} & \textit{DY}\\ \cline{2-4}
  & Opposition (Nay) & \textit{RN} & \textit{RN}\\ \hline
\end{tabular}
\caption{Speech classes}
\label{tab:table1}
\end{table}

\begin{table}[H]
\centering
\begin{tabular}{ | l | c | c | c | c | }
  \hline
  \textbf{} & \textbf{\# Speeches} & \textbf{\# Sentences} & \textbf{Avg. speech length} & \textbf{Avg. sentence length}\\ \hline
  RY & 1259 & 20697 & 16.4 sentences & 23.0 words\\ \hline
  RN & 151 & 3552 &	23.5 sentences & 23.5 words\\ \hline
  DY & 222 & 4342 &	19.6 sentences & 23.6 words\\ \hline
  DN & 1139 & 22280	& 19.6 sentences & 22.7 words\\ \hline \hline
  \textbf{Total} & \textbf{2771} & \textbf{50871} & \textbf{18.4 sentences} & \textbf{23.0 words}\\ \hline
\end{tabular}
\caption{Corpus overview}
\label{tab:table2}
\end{table}

\section{Method}
\label{sec:method}

\subsection{Language Model}
\label{sec:langmodel}
We use a simple statistical language model based on n-grams. In particular, we use 6-grams. That is, for each sequence of six consecutive words we calculate the probability of seeing the sixth word given the previous five ones. That allows us to determine very quickly all words which can occur after the previous five ones and how likely each of them is.

\subsection{Topic Model}
\label{sec:topicmodel}

For our topic model we use a Justeson and Katz (J\&K) POS tag filter for two- and three-word terms~\ref{itm:ref10}. As suggested by WordHoard~\ref{itm:ref23} we expanded the list of POS tag patterns by the sequence Noun-Conjunction-Noun. We determined the POS tags for each sentence in the corpus and identified then all two- and three-word terms that match one of the patterns. For the POS tagging we used maxent treebank pos tagging model from the Natural Language Toolkit (NLTK) for Python. It uses the maximum entropy model and was trained on the Wall Street Journal subset of the Penn Tree bank corpus~\ref{itm:ref24}.

Some of the terms are very generic and appear very often in all classes. In order to find those terms that appear particularly often in a certain class we calculate a significance score. Our significance score $Z$ is defined by the ratio of the probability of seeing a word $w$ in a certain class $c$ to the probability to see the word in the entire corpus:

\begin{equation*}
Z(w,c)=\frac{P(w | c)}{P(w)}
\end{equation*}

This significance score gives information about how often a term occurs in a certain class compared to the entire corpus. That is, every score greater than 1.0 indicates that in the given class a certain term occurs more often than average. We consider all phrases which occur at least 20 times in the corpus and have a ratio greater than 1. These terms represent the topics of the corpus. Table~\ref{tab:table3} lists the top ten topics of each class ordered by their score. All these terms represent meaningful topics and it seems reasonable that there were debates about them.

\begin{table}
\centering
\begin{tabular}{|>{\centering\arraybackslash}m{.215\textwidth}|>{\centering\arraybackslash}m{.215\textwidth}|>{\centering\arraybackslash}m{.215\textwidth}|>{\centering\arraybackslash}m{.215\textwidth}|}
  \hline
  \textbf{RY} & \textbf{RN} & \textbf{DY} & \textbf{DN} \\ \hline
  head start program & inner cell & cell research enhancement & school of law\\ \hline 
public law & inner cell mass & research enhancement act & cbc alternative\\ \hline 
death tax & human embryo & spinal cord & cbc budget\\ \hline 
budget request & human life & vitro fertilization & professor of law\\ \hline 
community protection act & adult stem cell & clean coal & republican budget\\ \hline 
community protection & world trade organization & stem cell & gun industry\\ \hline 
gang deterrence & adult stem & air national guard & big oil\\ \hline 
federal jurisdiction & world trade & human embryonic stem & judicial conference\\ \hline 
committee on homeland & sickle cell & stem cell research & democratic alternative\\ \hline 
deterrence and community & associate professor & heart disease & middle class\\ \hline \hline
\textbf{Total: 205 topics} & \textbf{Total: 84 topics} & \textbf{Total: 98 topics} & \textbf{Total: 182 topics}\\ \hline 
\end{tabular}
\caption{Top topics per class}
\label{tab:table3}
\end{table}

\subsection{Speech Generation}
\label{sec:speechgen}

For the speech generation one has to specify the desired class which consists of the political party and the intended vote. Based on the selected class the corresponding models for the generation are picked. 
From the language model of the selected class we obtain the probabilities for each 5-gram that starts a speech. From that distribution we pick one of the 5-grams at random and use it as the beginning of our opening sentence. 
Then the system starts to predict word after word until it predicts the token that indicates the end of the speech. In order to predict the next word we first determine what topics the so far generated speech is about. This is done by checking every topic-term if it appears in the speech. For every occurring term we calculate the topic coverage $TC$ in our speech. The topic coverage is an indicator of how well a certain topic $t$ is represented in a speech $S$. The following equation shows the definition of the topic coverage:

\begin{equation*}
    TC(S,t,c)=\frac{\# \text{ occurrences of t in S}}{\# \text{ occurrences of t in all speeches of class c}}
\end{equation*}

We rank all topics by their topic coverage values and pick the top 3 terms as our current topic set $T$. For these 3 terms we normalize the values of the ratios so that they sum up to 1. This gives us the probability $P(t | S,c)$ of seeing a topic $t$ in our current speech $S$ of class $c$.

The next step is to find our candidate words. All words which have been seen in the training data following the previous 5-gram are our candidates. For each candidate we calculate the probability of the language model $P_{language}$ and the probability of the topic model $P_{topic}$.
 
$P_{language}$ tells how likely this word is to occur after the previous 5 ones. This value can be directly obtained by the language model of the specified class. $P_{topic}$ tells how likely the word w is to occur in a speech which covers the current topics $T$. The following equation shows the definition of $P_{topic}$ where $X$ denotes our dataset and $X(c)$ is the subset containing only speeches of class $c$.

\begin{equation*}
P_{topic} (w | T,c)= \frac{\sum_{S' \in{X(c)}}[(\# \text{ occurrences of w in S'} )*\sum_{t \in T} P(t | S',c)] } {\sum_{S' \in X(c)}[(\# \text{ words in S'} )*\sum_{t \in T} P(t | S',c) ] * \varepsilon}
\end{equation*}

The factor $\varepsilon$ prevents divisions by zero is set to a very small value ($\varepsilon =0.001$). The probabilities for all candidate words are normalized so that they sum up to 1.

With the probabilities from the language model and the topic model we can now calculate the probability of predicting a certain word. This is done by combining those two probabilities. The weighting factor $\lambda$ balances the impact of the two probabilities. Furthermore, we want to make sure that a phrase is not repeated again and again. Thus, we check how often the phrase consisting of the previous five words and the current candidate word has already occurred in the generated speech and divide the combined probability by this value squared plus 1. So if this phrase has not been generated yet the denominator of this fraction is 1 and the original probability remains unchanged. The following equation shows how to calculate for a word $w$ the probability of being predicted as next word of the incomplete speech $S$:
\begin{equation*}
P_{word} (w | S)= \frac{\lambda P_{language}+(\lambda -1) P_{topic}}{1+ (\# \text{ occurrences of ($S[-5:]+w$) in S})^2  }
\end{equation*}

From the distribution given by the normalized probabilities of all candidate words we pick then one of the words at random. Then the whole procedure starts again with assessing the current topics. This is repeated until the end-of-speech token is generated or a certain word limit is reached.

Instead of using the probability distribution of the candidates we could have also just picked the word with the highest probability. But then the method would be deterministic. Using the distribution to pick a word at random enables the generator to produce every time a different speech.

\section{Alternative Methods}
\label{sec:altmethods}
In this section we present some alternative approaches which were pursued in the course of this project. These methods have not shown sufficiently good results and were therefore not further pursued.

\subsection{Recurrent Neural Networks}
Instead of using n-grams we also considered using Recurrent Neural Networks (RNN) as language models. Our approach was heavily based on the online tutorial from Denny Britz~\ref{itm:ref26}. The RNN takes as input a sequence of words and outputs the next word.  We limited the vocabulary to the 6000 most frequent words. Words were represented by one-hot-encoded feature vectors. The RNN had 50 hidden layers and used tanh as activation function. For assessing the error we used cross-entropy loss function. Furthermore we used Stochastic Gradient Descent (SGD) to minimize the loss and Backpropagation Through Time (BPTT) to calculate the gradients.

After training the network for 100 time epochs ($\sim$ 14 h) the results were still pretty bad. Most of the generated sentences were grammatically incorrect. There are many options to improve the performance of RNNs but due to the good performance shown by n-grams, the time-consuming training, and the limited time for this project we have decided to not further purse this approach.

\subsection{Latent Dirichlet Allocation}

As alternative to the J\&K POS tag filter we used LDA as topic model. In particular we used the approach from Lau et al.~\ref{itm:ref2}. That is, we removed all occurrences of stop words, stemmed the remaining words, replaced the 1000 most-frequent bigrams with single tokens, and deleted the 200 most frequent terms from the vocabulary before applying ordinary LDA. Since our dataset contains speech segments from 53 different debates we set the number of underlying topics to 53. Some of the results represented quite meaningful topics. However, the majority did not reveal any useful information. Table~\ref{tab:table4} shows some examples of good and bad results from LDA. It can be seen that the extracted terms of the bad examples are very generic and do not necessarily indicate a meaningful topic.

\begin{table}
\centering
\begin{tabularx}{\textwidth}{ | c | X | }
  \hline
  \textbf{Good examples} & \Centerstack[l]{-- iraq war us presid support vote administr congress \\ -- job make work compani busi right american good \\ -- economi job see need percent continu import now \\ -- program fund educ cut provid health help million \\ -- cut republican billion will pay percent benefit cost}\\ \hline
  \textbf{Bad examples} & \Centerstack[l]{-- issu countri address us today talk need can \\ -- go peopl get know want happen can say \\ -- go say talk want peopl get just said \\ --  member committe hous chang process standard vote \\ -- bill can mean use vote just first take}\\ \hline
\end{tabularx}
\caption{Results from LDA}
\label{tab:table4}
\end{table}

\subsection{Sentence-based approach}
For the speech generation task we have also pursued a sentence-based approach in the beginning of this project. The idea of the sentence-based approach is to take whole sentences from the training data and concatenate them in a meaningful way. We start by picking a speech of the desired class at random and take the first sentence of it. This will be the start sentence of our speech. Then we pick 20 speeches at random from the same class. We compare our first sentence with each sentence in those 20 speeches by calculating a similarity measure. The next sentence is than determined by the successor of the sentence with the highest similarity. In case no sentence shows sufficient similarity (similarity score below threshold) we just take the successor of our last sentence. In the next step we pick again 20 speeches at random and compare each sentence with the last one in order to find the most similar sentence. This will be repeated until we come across the speech-termination token or the generated speech reaches a certain length. 

The crucial part of this method is the measure of similarity between two sentences. Our similarity is composed of structural and textual similarity. Both are normalized to a range between 0 and 1 and weighted through a factor $\lambda$. We compute the similarity between two sentences $A$ and $B$ as follows:
\begin{equation*}
  Sim(A,B)=\lambda Sim_{struct} (A,B)+(1-\lambda)  Sim_{text} (A,B)  
\end{equation*}

For the structural similarity we compare the POS tags of both sentences and determine the longest sequence of congruent POS tags. The length of this sequence, normalized by the length of the shorter sentence, gives us the structural similarity. The structural similarity measure aims to support smooth sentence transitions. That is, if we find sentences which have a very similar sentence structure, it is very likely that they connect well to either of their following sentences. The textual similarity is defined by the number of trigrams that occur in both sentences, normalized by the length of the longer sentence. This similarity aims to find sentences which use the same words. 

The obvious advantage of the sentence-based approach is that every sentence is grammatically correct since they originate directly from the training data. However, connecting sentences reasonable is a very challenging task. A further step to improve this approach would be to extend the similarity measure by a topical similarity and a semantic similarity. The topical similarity should measure the topical correspondence of the originating speeches, while the semantic similarity should help to find sentences which express the same meaning although using different words. However, the results from the word-based approach were more promising and therefore we have decided to discard the sentence-based approach.

\section{Experiments}
\label{sec:experiments}
This section describes the experimental setup we used to evaluate our system. Furthermore, we present here two different approach of evaluating the quality of generated speeches.

\subsection{Setup}
In order to test our implemented methods we performed an experimental evaluation. In this experiment we generated ten speeches, five for class \textit{DN} and five for class \textit{RY}. We set the weighting factor $\lambda$ to 0.5 which means the topic and the language model have both equal impact on predicting the next word. The quality of the generated speeches was then evaluated. We used two different evaluation methods: a manual evaluation and an automatic evaluation. Both methods will be described in more detail in the following paragraphs of this section. The generated speeches can be found in the appendix of this report.

\subsection{Manual Evaluation}
For the manual evaluation we have defined a list of evaluation criteria. That is, a generated speech is evaluated by assessing each of the criterion and assigning a score between 0 and 3 to it. Table~\ref{tab:table5} lists all evaluation criteria and describes the meaning of the different scores.

\begin{table}
\centering
\begin{tabularx}{\textwidth}{ | m{.175\textwidth} | X | }
  \hline
  \multirow{2}{*}{\parbox{.15\textwidth}{\textbf{Grammatical correctness}}} & Are the sentences grammatically correct? \\ \cline{2-2}
  & \Centerstack[l]{0 ... The majority of the sentences are grammatically incorrect \\ 1 ... More than 50\% of the sentences contain mistakes \\ 2 ... Some sentences contain minor mistakes.  \\ 3 ... The majority of the sentences are grammatically correct} \\ \hline
  \multirow{2}{*}{\parbox{.15\textwidth}{\textbf{Sentence transitions}}} & How well do consecutive sentences connect? Reasonable references to previous sentences? \\ \cline{2-2}
  & \Centerstack[l]{0 ... Bad/no transitions, Incorrect use of references \\ 1 ... A few meaningful transitions, Partly incorrect use of references \\ 2 ... Most transitions and references are meaningful \\ 3 ... Very good transitions. Contains meaningful references to previous \\\ \hspace{.5cm} sentences.} \\ \hline
  \multirow{2}{*}{\textbf{Speech structure}} & Reasonable start / end of speech? Reasonable structuring of arguments (claim, warrant/evidence, conclusion)? Good flow? \\ \cline{2-2}
  & \Centerstack[l]{0 ... No clear structure. Random sentences. \\ 1 ... Contains some kind of structure. Arguments are very unclear.   \\ 2 ... Clear structure. Most of the arguments are meaningfully arranged. \\ 3 ... Very good structure. Contains meaningful start and end of speech. \\\ \hspace{.5cm} Clear breakdown of arguments.} \\ \hline
  \multirow{2}{*}{\textbf{Speech content}} & Same topic in entire speech? Reasonable arguments? \\ \cline{2-2}
  & \Centerstack[l]{0 ... No clear topic. Random arguments. \\ 1 ... Arguments cannot be assigned to a single topic. \\ 2 ... The majority of the speech deals with one topic. Most arguments are \\\ \hspace{.5cm} meaningful. \\ 3 ... Speech covers one major topic and contains meaningful arguments.} \\ \hline
\end{tabularx}
\caption{Evaluation criteria}
\label{tab:table5}
\end{table}

\subsection{Automatic Evaluation}
The automatic evaluation aims to evaluate both the grammatical correctness and the consistency of the speech in terms of its content. For evaluating the grammatical correctness we identify for each sentence of the speech its POS tags. Then we check all sentences of the entire corpus whether one has the same sequence of POS tags. Having a sentence with the same POS tag structure does not necessarily mean that the grammar is correct. Neither does the lack of finding a matching sentence imply the existence of an error. But it points in a certain direction. Furthermore, we let the system output the sentence for which it could not find a matching sentence so that we can evaluate those sentences manually.

In order to evaluate the content of the generated speech we determine the mixture of topics covered by the speech and order them by their topic coverage. That gives us information about the primary topic and secondary topics. Then we do the same for each speech in our dataset which is of the same class and compare the topic order with the one of the generated speech. We sum up the topic coverage values of each topic that occurs in both speeches at the same position. The highest achieved value is used as evaluation score. That is, finding a speech which covers the same topics with the same order of significance give us a score of 1.

\section{Results}
\label{sec:results}

In this section we present the results from our experiments. Table~\ref{tab:table6} shows the results from the manual evaluation. Note that each criterion scores between 0 and 3 which leads to a maximum total score of 12. The achieved total score range from 5 to 10 with an average of 8.1. In particular, the grammatical correctness and the sentence transitions were very good. Each of them scored on average 2.3 out of 3. The speech content yielded the lowest scores. This indicates that the topic model may need some improvement.

\begin{table}
\centering
\begin{tabularx}{\textwidth}{ | X | >{\centering\arraybackslash}m{.135\textwidth} | >{\centering\arraybackslash}m{.135\textwidth} |>{\centering\arraybackslash}m{.135\textwidth} |>{\centering\arraybackslash}m{.135\textwidth} || >{\centering\arraybackslash}m{.135\textwidth} | }
  \hline
   &\textbf{Grammatical correctness} & \textbf{Sentence transition} &  \textbf{Speech structure} & \textbf{Speech Content} & \textbf{Total} \\ \hline  
\textbf{DN\#1} & 2 & 2 & 3 & 2 & 9 \\ \hline 
\textbf{DN\#2} & 3 & 3 & 3 & 1 & 10 \\ \hline 
\textbf{DN\#3} & 2 & 2 & 1 & 1 & 6 \\ \hline 
\textbf{DN\#4} & 3 & 3 & 2 & 1 & 9 \\ \hline 
\textbf{DN\#5} & 3 & 3 & 2 & 1 & 9 \\ \hline 
\textbf{RY\#1} & 2 & 1 & 1 & 1 & 5 \\ \hline 
\textbf{RY\#2} & 2 & 2 & 1 & 1 & 6 \\ \hline 
\textbf{RY\#3} & 2 & 3 & 2 & 2 & 9 \\ \hline 
\textbf{RY\#4} & 2 & 2 & 3 & 3 & 10 \\ \hline 
\textbf{RY\#5} & 2 & 2 & 2 & 2 & 8 \\ \hline 
\textit{\textbf{Average}} & \textit{2.3} & \textit{2.3} & \textit{2} & \textit{1.5} & \textit{8.1} \\ \hline 
\end{tabularx}
\caption{Results from manual evaluation}
\label{tab:table6}
\end{table}

Table~\ref{tab:table7} shows the results from the automatic evaluation. The automatic evaluation confirms pretty much the results from the manual evaluation. Most of the speeches which achieved a high score in the manual evaluation scored also high in the automatic evaluation. Furthermore, it also confirms that the overall the grammatical correctness of the speeches is very good while the content is a bit behind.

\begin{table}
\centering
\begin{tabularx}{\textwidth}{ | X | >{\centering\arraybackslash}m{.25\textwidth} | >{\centering\arraybackslash}m{.25\textwidth} || >{\centering\arraybackslash}m{.25\textwidth} | }
  \hline
   &\textbf{Grammatical correctness} & \textbf{Speech content} & \textbf{Mean} \\ \hline  
\textbf{DN\#1} & 0.65 & 0.49 & 0.57 \\ \hline 
\textbf{DN\#2} & 0.5 & 0.58 & 0.54 \\ \hline 
\textbf{DN\#3} & 0.86 & 0.25 & 0.56 \\ \hline 
\textbf{DN\#4} & 0.61 & 0.34 & 0.48 \\ \hline 
\textbf{DN\#5} & 0.70 & 0.25 & 0.48 \\ \hline 
\textbf{RY\#1} & 0.52 & 0.28 & 0.4 \\ \hline 
\textbf{RY\#2} & 0.65 & 0.68 & 0.67 \\ \hline 
\textbf{RY\#3} & 0.25 & 0.95 & 0.6 \\ \hline 
\textbf{RY\#4} & 0.63 & 0.8 & 0.72 \\ \hline 
\textbf{RY\#5} & 0.5 & 0.21 & 0.36 \\ \hline 
\textit{\textbf{Average}} & \textit{0.59} & \textit{0.48} & \textit{0.54} \\ \hline 
\end{tabularx}
\caption{Results from automatic evaluation}
\label{tab:table7}
\end{table}

\section{Conclusion}
\label{sec:conclusion}
In this report we have presented a novel approach of training a system on speech transcripts in order to generate new speeches. We have shown that n-grams and J\&K POS tag filter are very effective as language and topic model for this task. We have shown how to combine these models to a system that produces good results. Furthermore, we have presented different methods to evaluate the quality of generated texts. In an experimental evaluation our system performed very well. In particular, the grammatical correctness and the sentence transitions of most speeches were very good. However, there are no comparable systems which would allow a direct comparison. 

Despite the good results it is very unlikely that these methods will be actually used to generate speeches for politicians. However, the approach applies to the generation of all kind of texts given a suitable dataset. With some modifications it would be possible to use the system to summarize texts about the same topic from different source, for example when several newspapers report about the same event. Terms that occur in the report of every newspaper would get a high probability to be generated.

All of our source code is available on GitHub~\ref{itm:ref_source}. We explicitly encourage others to try using, modifying and extending it. Feedback and ideas for improvement are most welcome.

\section*{References}
\begin{enumerate}[label={[\arabic*]},leftmargin=*]
\item \label{itm:ref1} D. Blei, A. Ng, M. Jordan. (2003). Latent Dirichlet allocation. Journal of Machine Learning Research, 3:993–1022.
\item \label{itm:ref2} Lau, J. H., Baldwin, T., Newman, D. (2013). On collocations and topic models. ACM Transactions on Speech and Language Processing (TSLP), 10(3), 10.
\item \label{itm:ref3} Fürnkranz, J. (1998). A study using n-gram features for text categorization. Austrian Research Institute for Artifical Intelligence, 3(1998), 1-10.
\item \label{itm:ref4} Cavnar, W. B., Trenkle, J. M. (1994). N-gram-based text categorization. Ann Arbor MI, 48113(2), 161-175.
\item \label{itm:ref5} Smadja, F. A. (1991, June). From n-grams to collocations: An evaluation of Xtract. In Proceedings of the 29th annual meeting on Association for Computational Linguistics (pp. 279-284). Association for Computational Linguistics.
\item \label{itm:ref6} Abney, S. (1990, October). Rapid incremental parsing with repair. In Proceedings of the 6th New OED Conference: Electronic Text Research (pp. 1-9).
\item \label{itm:ref7} Wang, X., McCallum, A., Wei, X. (2007, October). Topical n-grams: Phrase and topic discovery, with an application to information retrieval. In Data Mining, 2007. ICDM 2007. Seventh IEEE International Conference on (pp. 697-702). IEEE.
\item \label{itm:ref8} Wallach, H. M. (2006, June). Topic modeling: beyond bag-of-words. In Proceedings of the 23rd international conference on Machine learning (pp. 977-984). ACM.
\item \label{itm:ref9}Griffiths, T. L., Steyvers, M., Tenenbaum, J. B. (2007). Topics in semantic representation. Psychological review, 114(2), 211.
\item \label{itm:ref10} Justeson, J. S., Katz, S. M. (1995). Technical terminology: some linguistic properties and an algorithm for identification in text. Natural language engineering, 1(01), 9-27.
\item \label{itm:ref11} Wacholder, N. (1998). Simplex NPs clustered by head: a method for identifying significant topics within a document. In The Computational Treatment of Nominals: Proceedings of the Workshop (pp. 70-79).
\item \label{itm:ref12} Blei, D. M., Lafferty, J. D. (2007). A correlated topic model of science. The Annals of Applied Statistics, 17-35.
\item \label{itm:ref13} Reiter, E., Dale, R. (1997). Building applied natural language generation systems. Natural Language Engineering, 3(1), 57-87.
\item \label{itm:ref14} Reiter, E., Sripada, S., Hunter, J., Yu, J., Davy, I. (2005). Choosing words in computer-generated weather forecasts. Artificial Intelligence, 167(1), 137-169.
\item \label{itm:ref15} Lavoie, B., Rambow, O., Reiter, E. (1996, June). The modelexplainer. In Proceedings of the 8th international workshop on natural language generation (pp. 9-12).
\item \label{itm:ref16} Caldwell, D. E., Korelsky, T. (1994, October). Bilingual generation of job descriptions from quasi-conceptual forms. In Proceedings of the fourth conference on Applied natural language processing (pp. 1-6). Association for Computational Linguistics.
\item \label{itm:ref17} Paris, C., Vander Linden, K., Fischer, M., Hartley, A., Pemberton, L., Power, R., Scott, D. (1995, August). A support tool for writing multilingual instructions. In INTERNATIONAL JOINT CONFERENCE ON ARTIFICIAL INTELLIGENCE (Vol. 14, pp. 1398-1404). LAWRENCE ERLBAUM ASSOCIATES LTD.
\item \label{itm:ref18} Matt Thomas, Bo Pang, Lillian Lee. 2006. Get out the vote: determining support or opposition from congressional floor-debate transcripts. In Proceedings of the 2006 Conference on Empirical Methods in Natural Language Processing (EMNLP '06).
\item \label{itm:ref19} Iyyer, M., Enns, P., Boyd-Graber, J., Resnik, P. (2014). Political ideology detection using recursive neural networks. In Association for Computational Linguistics.
\item \label{itm:ref20} Bradley, M., Lang, P.: Affective norms for english words (ANEW): Stimuli, instruction manual and affective ratings. Technical report C-1. The Center for Research in Psychophysiology, University of Florida (1999).
\item \label{itm:ref21} Gross, J., Acree, B., Sim, Y., Smith, N. A. (2013). Testing the Etch-a-Sketch Hypothesis: A Computational Analysis of Mitt Romney's Ideological Makeover During the 2012 Primary vs. General Elections. In APSA 2013 Annual Meeting Paper.
\item \label{itm:ref22} SCIgen - An Automatic CS Paper Generator. https://pdos.csail.mit.edu/archive/scigen/. 
Accessed: 2015-12-12
\item \label{itm:ref23} Finding Multiword Units. http://wordhoard.northwestern.edu/userman/analysis-multiwordunits.html. Accessed: 2015-12-12
\item \label{itm:ref24} The Penn Treebank Project. https://www.cis.upenn.edu/~treebank/. Accessed: 2015-12-12
\item \label{itm:ref25} Congressional speech data. http://www.cs.cornell.edu/home/llee/data/convote.html. 
Accessed: 2015-12-12
\item \label{itm:ref26} Recurrent Neural Networks Tutorial, Part 2 – Implementing a RNN with Python, Numpy and Theano. http://www.wildml.com/2015/09/recurrent-neural-networks-tutorial-part-2-implementing-a-language-model-rnn-with-python-numpy-and-theano/. Accessed: 2015-12-12
\item \label{itm:ref_source} Valentin Kassarnig. Political Speech Generator repository at https://github.com/valentin012/conspeech
\end{enumerate}


\newpage

\section*{Appendix}

\subsection*{Generated speeches from experiment}

\subsubsection*{DN\#1}
\_\_START\_\_ mr. speaker , i thank my colleague on the committee on rules .
i rise in full support of this resolution and urge my colleagues to support this bill and urge my colleagues to support the bill .
mr. speaker , supporting this rule and supporting this bill is good for small business .
it is great for american small business , for main street , for jobs creation .
we have an economy that has created nearly 2 million jobs in the past couple of months : apparel , textiles , transportation and equipment , electronic components and equipment , chemicals , industrial and commercial equipment and computers , instruments , photographic equipment , metals , food , wood and wood products .
virtually every state in the union can claim at least one of these industrial sectors .
in fact , one young girl , lucy , wanted to make sure that the economy keeps growing .
that should not be done on borrowed money , on borrowed time .
it should be done with a growing economy .
it is under this restraint , with this discipline , that this budget comes before the house , and we should honor that work .
\_\_END\_\_

\subsubsection*{DN\#2}
\_\_START\_\_ mr. speaker , for years , honest but unfortunate consumers have had the ability to plead their case to come under bankruptcy protection and have their reasonable and valid debts discharged .
the way the system is supposed to work , the bankruptcy court evaluates various factors including income , assets and debt to determine what debts can be paid and how consumers can get back on their feet .
stand up for growth and opportunity .
pass this legislation .
\_\_END\_\_

\subsubsection*{DN\#3}
\_\_START\_\_ mr. speaker , i yield back the balance of my time , and i want to commend , finally , the chairman of the committee , there will be vigorous oversight of the department of justice on a regular and on a timely basis , and the answer to how many civil liberties violations have been proven is none .
repeatedly they have said there are no civil liberties violations that the inspector general has been able to uncover .
further , i resisted a premature repeal or extension of the sunset prior to this congress because i felt it was important that the oversight be done for as long a time as possible so that the congress will be able to vote and a decision can be made today .
mr. speaker , i reserve the balance of my time , and i want to thank the gentleman from texas for helping put together this package and for all the work that he and his staff put into this bill .
this was an important thing for us to go through , and i think that we produced a good bill at the end of that dark ally over there .
and the gentleman says : because there is more light over here .
sometimes i think the way we look at these medical issues , instead of looking at the cost savings involved with prevention , we simply are able to look at how much it saves in the long run .
again , i look at such things as if we are able to have more people go to federally approved health centers , community health centers in their community instead of showing up in the emergency departments , yes , it may cost money ; the president called for a couple billion dollars to put into those community health centers .
but if it is going to relate to state law , that is the discussion that needs to take place .
my state may have lucked out because a clerical error in this particular case did not refer specifically to the utah state law ; and , therefore , it may not be applicable .
but the fear factor is still there , that in the future he continue that policy .
\_\_END\_\_

\subsubsection*{DN\#4}
\_\_START\_\_ mr. speaker , for years , honest but unfortunate consumers have had the ability to plead their case to come under bankruptcy protection and have their reasonable and valid debts discharged .
the way the system is supposed to work , the bankruptcy court evaluates various factors including income , assets and debt to determine what debts can be paid and how consumers can get back on their feet , they need to have money to pay for child care .
they need transportation .
it allows them to get reestablished , and we think this is certainly very helpful .
and then it also allows faith-based organizations to offer job training service .
we think this is critical and has great potential .
at the present time , brazil mandates 23 percent of their fuel supply be from ethanol .
we certainly could hit 7 or 8 percent in this country .
mr. speaker , this is a very modest proposal .
i think it is important that this resolution be considered quickly , so that members may be appointed to the task force and can begin their work and produce a report by june 2006 .
\_\_END\_\_

\subsubsection*{DN\#5}
\_\_START\_\_ mr. speaker , i yield myself the time remaining .
mr. speaker , i rise today in support of the rule on h.r. 418 .
our nation's immigration policy has been of top concern in recent years , and for good reason .
with between eight and twelve million illegal aliens in the united states , the late ronald wilson reagan , enshrined these three words as part of american policy : trust but verify .
the legislation on the floor today deals with verification .
i say as one who opposed a trading agreement with china that this legislation brings the monitoring capacity necessary to understand what happens in international trade .
simply stated , madam speaker , if you want to cut those things , you can put it in your program .
if you do not like that , you better go out and lobby against what they are doing in in vitro fertilization clinics throughout the u.s. , about 2 percent are discarded annually -- that is about 8 , 000 -- 11 , 000 embryos that could be slated for research .
allowing the option of donating these excess embryos to research is similar to donating organs for organ transplantation in order to save or improve the quality of another person's life .
the bottom line is that class-action reform is badly needed .
currently , crafty lawyers are able to game the system by filing large , nationwide class-action suits in certain preferred state courts such as madison county , illinois , where judges are quick to certify classes and quick to approve settlements that give the lawyers millions of dollars in fees .
this problem will be addressed by providing greater scrutiny over settlements that involve coupons or very small cash amounts .
this legislation also ensures that deserving plaintiffs are able to make full use of the class action system .
it allows easier removal of class action cases to federal courts .
this is important because class actions tend to affect numerous americans and often involve millions of dollars .
federal court is the right place for such large lawsuits .
moving more class actions to federal courts also prevents one of the worst problems in class actions today , forum shopping .
mr. speaker , while many concessions were made on both sides , this is still a very worthwhile bill that contains many good reforms , and i fully support it and look forward to its enactment into law and also encourage my colleagues to support this bill .
\_\_END\_\_

\subsubsection*{RY\#1}
\_\_START\_\_ 
mr. speaker , i yield 2 minutes to the gentleman from illinois ( mr. hyde ) , my dear friend , with whom i agree on some things but not on this issue , although the majority of the bill i know is consistent with the gentleman from california's ( mr. lantos ) and the gentleman from virginia with their very wise substitute give a chance to help the consumer and declare energy independence .
i also want to point out that this bill is far from perfect .
in many respects it is troubling .
this congress has a proven history of lax oversight of the administration , and there is a difference .
\_\_END\_\_

\subsubsection*{RY\#2}
\_\_START\_\_ 
mr. speaker , the gentleman is absolutely right .
the amazing thing to me when i was listening to the republicans in the last hour is when they were trying to make the analogy to their households and talking about their kids .
and one of the most significant broken promises is in the area of making higher educational opportunities more available to minority and low-income students .
i am so proud of the fact that every iraqi school child on the opening day of school had received a book bag with the seal of the u.s. , pencils , pads , all kinds of things , free of charge .
i had just come back from iraq , and they had been there on the first day of this new congress , the republican majority is publicly demonstrating what has been evident for some time , and that is its arrogance , its pettiness , its shortsighted focus on their political life rather than to decide how we are each of us fit to govern .
here is the thing .
we have this rules package before us .
they did some flash last night so that the press is saying , oh , they blinked .
they did blink on a couple of different scores , but the fundamental challenge to the ethical standard of the house being enforced is still in this rules package are destructive , and they are unethical .
mr. speaker , i reserve the balance of my time .
mr. chairman , this bill frightens me .
it scares me .
i would hope that we could deal with this in as bipartisan a fashion as possible so that when we send it to the other body that we may have more success there , more success out of conference , and send a bill to the president that will facilitate both energy independence and the effective and efficient discovery , development , and delivery at retail to the consumer of energy options .
i do not know if politics was part of that .
maybe someone can answer that question .
but therein lies the problem , that from time to time need to be recognized .
that is what this is about .
this bill is opposed by every consumer group , by all the bankruptcy judges , the trustees , law professors , by all of organized labor , by the military groups , by the civil rights organizations , and by every major group concerned about seniors , women , and children are dead ; the fact that hundreds of thousands more have become evacuees in the richest country in the world .
our children will then be forced to live with the consequences of an undereducated workforce , a weak economy , and a society where good health and social justice are only afforded to the most privileged .
mr. speaker , i reserve the balance of my time to read the resolution that i believe ought to be before us , mr. speaker .
the president has a credibility gap when it comes to iraq .
we have been misled too often , and it is time to go back and revisit those. '' i would remind the house that it was widely pointed out when that legislation was before us what a remarkable example of bipartisanship and legislative cooperation it was .
of course , the defense appropriations bill is of great interest to our members .
\_\_END\_\_

\subsubsection*{RY\#3}
\_\_START\_\_ 
mr. speaker , i rise today in opposition to the labor , health and human services and education appropriations conference report before us .
one month ago , the house of representatives voted this bill down because it failed to address the priorities of the american people : good jobs , safe communities , quality education , and access to health care .
with over 7 million americans out of work .
yet the bill cuts \$ 437 million out of training and employment services .
that is the lowest level of adult training grants in a decade .
this bill also cuts the community college initiative , the president's initiative for community colleges , an effort to train workers for high-skill , high-paying jobs .
it cuts that effort by $ 125 million and rescinds $ 125 million from funds provided last year , denying the help that the president was talking about giving to 100 , 000 americans of a continued education to help them get a new job .
this bill also cuts job search assistance through the employment service by 11 percent and cut state unemployment insurance and employment service offices are cut \$ 245 million eliminating help for 1.9 million people .
this bill is no better for those attending college full-time .
despite the fact that college costs have increased by \$ 3 , 095 , 34 percent , since 2001 .
consumers are expected to pay 52 percent more for natural gas , 30 percent more for home heating oil , you are expected to pay three times as much as you did 4 years ago , the first year president bush took office .
winter is around the corner , and so are skyrocketing increases in home heating costs .
families who heat with natural gas could see their fuel costs increase more than 70 percent in some parts of the country .
this honorable response to the tragedy of september 11 puts to shame what has been proposed today in the wake of hurricane katrina , that the workers in the afflicted area who are trying to put that area back together are not even going to be allowed to get a decent prevailing wage that they would otherwise be guaranteed under davis-bacon .
and yet while it is chiseling on the wages of those workers , it is bad for those countries that desperately need a middle class , it is bad for those workers , it is saying to the persons who make over \$ 400 , 000 a year , and we roll back cuts on the top 2 percent of americans , and by doing so , we have saved almost \$ 47 billion that we have used to invest in the human assets of this country , the american people .
\_\_END\_\_

\subsubsection*{RY\#4}

\_\_START\_\_ 
mr. speaker , i yield 2 minutes to the gentlewoman from california ( mrs. capps ) pointed out , after the knowledge was available and was continued to pursue the use of this compound as an additive to the fuels of our automobiles .
those communities now are stuck with the costs of either cleaning up that drinking water supply , finding an alternative source and dealing with it , and they must do so .
to suggest now that we are going to be giving to seniors , to keep them in nursing homes with alzheimer's and with parkinson's disease , just keep cutting it .
give more tax breaks to the richest one-tenth of 1 percent .
they call it the death tax .
i think that is a flaw in the bill .
that leads to the second point .
the bill specifically mentions weight gain and obesity .
well , i think most of us have a sense of what obesity is .
weight gain is a whole different issue , and weight gain may occur not from obesity , not from getting fat , not from putting on too many calories ; weight gain can occur for a variety of medical reasons related to a variety of different causes .
for example , i mean probably all of us have had a mom or a grandmom or an uncle to whom we say , hey , i noticed your legs are swelling again .
fluid retention .
fluid retention .
now , that can be from a variety of causes .
that is not from increased caloric intake .
that could have been , for example , from a food additive , maybe a cause that was not known to the public of some kind of additive in something that they had eaten or drank .
it may have been something that interfered with one of their medications and led to fluid retention .
i am just making up hypotheticals here .
or , the hypothetical , perhaps you have something that is actually a heart poison from some food additive that has no calories in it , zero calories in it , but over a period of time does bad things to the ability of under this bill , which i believe is absolutely essential for our health system .
at a time when our country has been severely impacted by natural disasters , it is extremely urgent that congress maintain csbg funding at its current level so that the delivery of much needed services to low-income people is not disrupted .
we have a responsibility to protect our environment -- as well as the diverse forms of life that share it .
the bipartisan substitute will help us achieve the goal .
i urge my colleagues on both sides of the aisle to protect the benefits that our constituents earned and deserve and to prevent the increase in the number of frivolous filings .
\_\_END\_\_

\subsubsection*{RY\#5}
\_\_START\_\_ 
mr. speaker , i yield 2 minutes to the gentlewoman from texas ( ms. jackson-lee ) , the gentleman from new jersey ( mr. andrews ) , for the leadership he has shown on this issue .
here we are again , mr. speaker .
year after year after year trying to get into federal court .
what it also does is minimizes the opportunity of those who can secure their local lawyer to get them into a state court and burdens them with the responsibility of finding some high-priced counsel that they can not afford to buy food .
seven million more people , an increase of 12 percent , and what does this combination of reconciliation in order to give tax cuts to people making more than \$ 500 , 000 .
footnote right there .
what about the committees of jurisdiction already in existence in congress .
and what about creating a circus atmosphere that drains resources from this congress do you not understand .
shamefully , the house will not have an opportunity to vote on the hastings-menendez independent katrina commission legislation , because republicans have blocked us from offering it .
just as they always do , republicans block what they can not defeat .
despite what republicans will suggest , today's debate is not about politics .
it is about the need for truth to assure the american people that we will not allow their retirement checks to be slashed to pay for private accounts .
it is time for congress , as part of the national marine sanctuary program , but there have been no hearings on this bill or any other bill to protect our oceans .
let us reject this unnecessary task force and get down to some real work .
mr. speaker , i reserve the balance of my time to the gentleman from maryland ( mr. cardin ) , who is the ranking member , was part and parcel of that , as well as the gentleman from virginia ( chairman tom davis ) is trying to do to improve the integrity of driver's licenses , but i find it interesting that the state of utah , while the gentleman from utah ( mr. bishop ) is arguing that they are not getting enough money for education , the state of utah legislature passed measures saying they do not want any kind of investigation of themselves .
the republicans control the white house , they control the senate , and they control the house of representatives .
mr. speaker , is it possible for us to let this young woman take her leave in peace .
\_\_END\_\_

\end{document}